\documentclass[runningheads]{llncs}
\usepackage[T1]{fontenc}
\usepackage{graphicx}
\usepackage{multirow}
\usepackage{amsmath}  
\usepackage{amssymb}  
\usepackage{caption}
\usepackage{subcaption}
\usepackage[T1]{fontenc}
\usepackage{xcolor}
\usepackage{hyperref}

\begin{document}
\title{Exploring Pre-training Across Domains for Few-Shot Surgical Skill Assessment}
%
%
\author{
Dimitrios Anastasiou\inst{1,2} \and
Razvan Caramalau\inst{3,5} \and
Nazir Sirajudeen\inst{1,8} \and
Matthew Boal\inst{4,8} \and 
Philip Edwards\inst{1,5} \and
Justin Collins\inst{6} \and
John Kelly\inst{6} \and
Ashwin Sridhar\inst{6} \and
Maxine Tran\inst{7} \and 
Faiz Mumtaz\inst{7} \and 
Nevil Pavithran\inst{7} \and 
Nader Francis\inst{8} \and 
Danail Stoyanov\inst{1,5} \and 
Evangelos B. Mazomenos\inst{1,2}}
\authorrunning{Dimitrios Anastasiou et al.}
%
\institute{UCL Hawkes Institute, University College London, UK \and
Dept of Medical Physics \& Biomedical Engineering, University College London, UK \and
Medtronic, Digital Technologies, UK \and
Gloucestershire Hospitals NHS Foundation Trust, UK \and
Dept of Computer Science, University College London, UK \and
University College London Hospitals NHS Foundation Trust, UK \and
Royal Free Hospital NHS Foundation Trust, UK \and
The Griffin Institute, UK
\\
\email{dimitrios.anastasiou.21@ucl.ac.uk}}
\maketitle              
\begin{abstract}
Automated surgical skill assessment (SSA) is a central task in surgical computer vision. Developing robust SSA models is challenging due to the scarcity of skill annotations, which are time-consuming to produce and require expert consensus. Few-shot learning (FSL) offers a scalable alternative enabling model development with minimal supervision, though its success critically depends on effective pre-training. While widely studied for several surgical downstream tasks, pre-training has remained largely unexplored in SSA. In this work, we formulate SSA as a few-shot task and investigate how self-supervised pre-training strategies affect downstream few-shot SSA performance. We annotate a publicly available robotic surgery dataset with Objective Structured Assessment of Technical Skill (OSATS) scores, and evaluate various pre-training sources across three few-shot settings. We quantify domain similarity and analyze how domain gap and the inclusion of procedure-specific data into pre-training influence transferability. Our results show that small but domain-relevant datasets can outperform large-scale, less aligned ones, achieving accuracies of 60.16\%, 66.03\%, and 73.65\% in the 1-, 2-, and 5-shot settings, respectively. Moreover, incorporating procedure-specific data into pre-training with a domain-relevant external dataset significantly boosts downstream performance, with an average gain of +1.22\% in accuracy and +2.28\% in F1-score; however, applying the same strategy with less similar but large-scale sources can instead lead to performance degradation. Code and models are available at \href{https://github.com/anastadimi/ssa-fsl}{\texttt{ssa-fsl}}.

\end{abstract}
\section{Introduction}
Automated surgical skill assessment (SSA) is a fundamental task in surgical data science. It offers substantial clinical value by enabling objective and scalable evaluation of technical performance, which is linked to patient safety and clinical outcomes \cite{Maier-Hein2017}. However, a major barrier to advancing automated SSA is the limited availability of annotated data. SSA annotations are typically based on frameworks such as the Objective Structured Assessment of Technical Skill (OSATS) \cite{Martin1997}, which assess surgical performance using Likert-scale scores across multiple skill domains. This process is time-consuming and requires expert consensus, often involving multiple annotators and significant effort per video \cite{Sirajudeen2024}. Moreover, most existing SSA models are trained in a task-specific manner, which limits their ability to generalize across different procedures, surgical tasks, or clinical settings \cite{Anastasiou2023,Li2022,Liu2021,Funke2019}. Consequently, deploying these models in new contexts often requires collecting a new set of skill annotations for training. As a result, scaling such models is particularly challenging given the annotation burden. This challenge underscores a core data engineering problem: how to effectively leverage limited annotated data for model development. Few-shot learning (FSL), which aims to learn from only a handful of labeled examples \cite{Parnami2022}, presents a promising and more scalable alternative for SSA. However, despite its potential, FSL for SSA remains significantly underexplored \cite{yanik2024}, leaving its broader applicability largely unexamined. In this work, we propose formulating SSA as a few-shot learning task and provide preliminary results toward establishing its feasibility and effectiveness on real-world surgical data.

While pre-training is crucial in FSL for learning transferable representations \cite{Parnami2022}, its effectiveness for few-shot SSA remains unexplored. Pre-training has become the standard approach in computer vision due to its ability to significantly reduce the amount of labeled data required to achieve strong downstream performance. Initially dominated by supervised methods using large-scale datasets (\textit{e.g.,} ImageNet \cite{Deng2009}), recent advances in self-supervised learning (SSL) have matched or surpassed these approaches by learning rich representations without annotations \cite{Wang2023,Radford2021,Caron2021}. This is particularly beneficial for surgical computer vision, where large annotated datasets are scarce, but abundant unlabeled surgical videos enable effective pre-training \cite{MaierHein2022}. Consequently, domain-specific pre-training becomes especially compelling, as SSL in surgical contexts has been shown to outperform generic pre-training on standard vision datasets \cite{Hirsch2023,Jaspers2025,Batic2024}. From a data engineering standpoint, SSL enables the efficient utilization of vast amounts of unlabeled surgical data, reducing reliance on expert annotations while preserving domain relevance.

One of the earliest benchmarks for pre-training in the surgical domain systematically evaluated several SSL methods, highlighting the impact of model parameterization and training data size, and, generalization across data from different institutions \cite{Ramesh2023}. Subsequent studies have demonstrated the effectiveness of large-scale SSL pre-training on both public and private surgical datasets \cite{Alapatt2024,Batic2023,Hirsch2023,Wang2023pretrain}, addressing questions related to pre-training dataset composition and selection, among others. More recently, Jaspers et al. \cite{Jaspers2025} explored the role of dataset diversity in SSL, finding that increased diversity in pre-training surgical data improves downstream performance, although their study was limited to segmentation tasks. While these works span a range of downstream tasks, from surgical phase recognition to tool detection, none have investigated the effects of SSL pre-training on the downstream task of SSA. Unlike other surgical tasks, SSA involves holistic reasoning over entire procedures, making it more sensitive to the quality of learned representations \cite{Li2022}. Moreover, the role of the domain gap, \textit{i.e.,} the similarity between the pre-training and downstream datasets, remains unexplored, particularly in terms of quantifying it and assessing its impact on downstream performance. In this work, we explore SSL pre-training strategies for few-shot SSA for the first time. We systematically evaluate the impact of pre-training data choices, and propose a framework to quantify the domain gap and assess its effect on downstream SSA performance.

In summary, the key contributions of this work are:

\begin{enumerate}
    \item We propose formulating SSA as a few-shot learning task and provide a systematic assessment of its feasibility and effectiveness on real surgical data.
    \item We evaluate the performance capabilities of few-shot SSA under various pre-training strategies, highlighting how different choices affect model performance.
    \item We explore how the domain gap between pre-training and downstream datasets, as well as the inclusion of procedure-specific data, impacts downstream SSA performance in low-data regimes, addressing key data engineering questions around dataset selection and structuring.
\end{enumerate}

\section{Methods}
\subsection{Downstream Surgical Skill Assessment}

\subsubsection{Dataset}
To evaluate few-shot performance for SSA, we use the publicly available SAR-RARP50 dataset \cite{Psychogyios2023}, which we further annotate with OSATS scores \cite{Martin1997}. SAR-RARP50 consists of 54 videos of robot-assisted radical prostatectomies, focusing on the suturing phase of the dorsal vascular complex. The procedures were performed by eight surgeons of varying experience levels, including consultants, senior registrars, and junior registrars. For our study, we selected 33 videos for manual annotation by two experts in robotic surgery using the OSATS scale with six skill categories: respect for tissue, suture/needle handling, time and motion, flow of operation, overall performance, and quality of final product \cite{Sirajudeen2024}. Each category is scored from 1 to 5, with the Global Rating Score (GRS) computed as the sum of all scores. We use the 33 annotated videos, denoted as SAR-RARP50\textsubscript{L}, for downstream evaluation. The remaining 21 unlabeled videos are used for pre-training and referred to as SAR-RARP50\textsubscript{U}.

\subsubsection{Task}
As these procedures are real clinical cases performed by trained surgeons, most GRS values naturally fall in the upper range, resulting in a narrow range between 19 and 30. To simplify the SSA task, we convert the GRS regression problem into a binary classification task by categorizing performance levels into two classes: proficient (GRS 19–24) and expert (GRS 25–30). 
This thresholding approach helps create a more balanced label distribution, which is beneficial for both model training and evaluation stability. A multi-class formulation is not feasible in this setup, as it would result in too few samples per class, making training and evaluation statistically unreliable. Prior work suggests that, to achieve sufficient statistical power with a moderate effect size and 80\% power, approximately 32 samples per class are needed for binary classification \cite{vanBelle2008}, and this requirement increases quadratically with the number of classes. Although our dataset includes fewer than 32 samples per class, this is acceptable for exploratory analysis and proof-of-concept purposes, particularly in the context of SSA, where data collection and annotation are resource-intensive.

Finally, the few-shot classification task is evaluated through a series of episodes, where each episode involves \textit{k}-shots (\textit{k} labeled samples) per class for training, with the remaining samples used for evaluation.

\subsection{Pre-training Datasets}

We selected our pre-training datasets based on their visual and semantic similarity to the downstream dataset, with the goal of investigating how domain gap influences few-shot SSA performance. Our pre-training datasets include SAR-RARP50\textsubscript{U}, RALPN, SurgToolLoc \cite{Zia2023}, and Something-Something-v2 \cite{Goyal2017}.

\noindent \textbf{RALPN} is a private dataset comprising 11 robot-assisted laparoscopic partial nephrectomy procedures, recorded at the Royal Free Hospital, London, UK. It contains clips from the suturing phase following tissue dissection on the kidney. Although relatively small in scale ($\sim$3.5 hours of footage), it was included due to its high procedural and visual similarity to SAR-RARP50, as both datasets involve real robotic surgery cases and contain suturing tasks.

\noindent \textbf{SurgToolLoc} is a publicly available dataset consisting of video clips from surgical training exercises conducted on a porcine model using the da Vinci Robotic Surgical System. It was selected because of its relevance to robotic surgery and its large scale ($\sim$206 hours of footage).

\noindent \textbf{Something-Something-v2} is a general-purpose action recognition dataset composed of short clips depicting human-object interactions. This dataset was included as a generic initialization baseline due to its large size ($\sim$232 hours of footage) and the hypothesis that the diversity of motion and manipulation patterns may still provide transferable representations useful for modeling surgical tasks.

\begin{figure*}[t]
    \centering
    \includegraphics[width=1\textwidth]{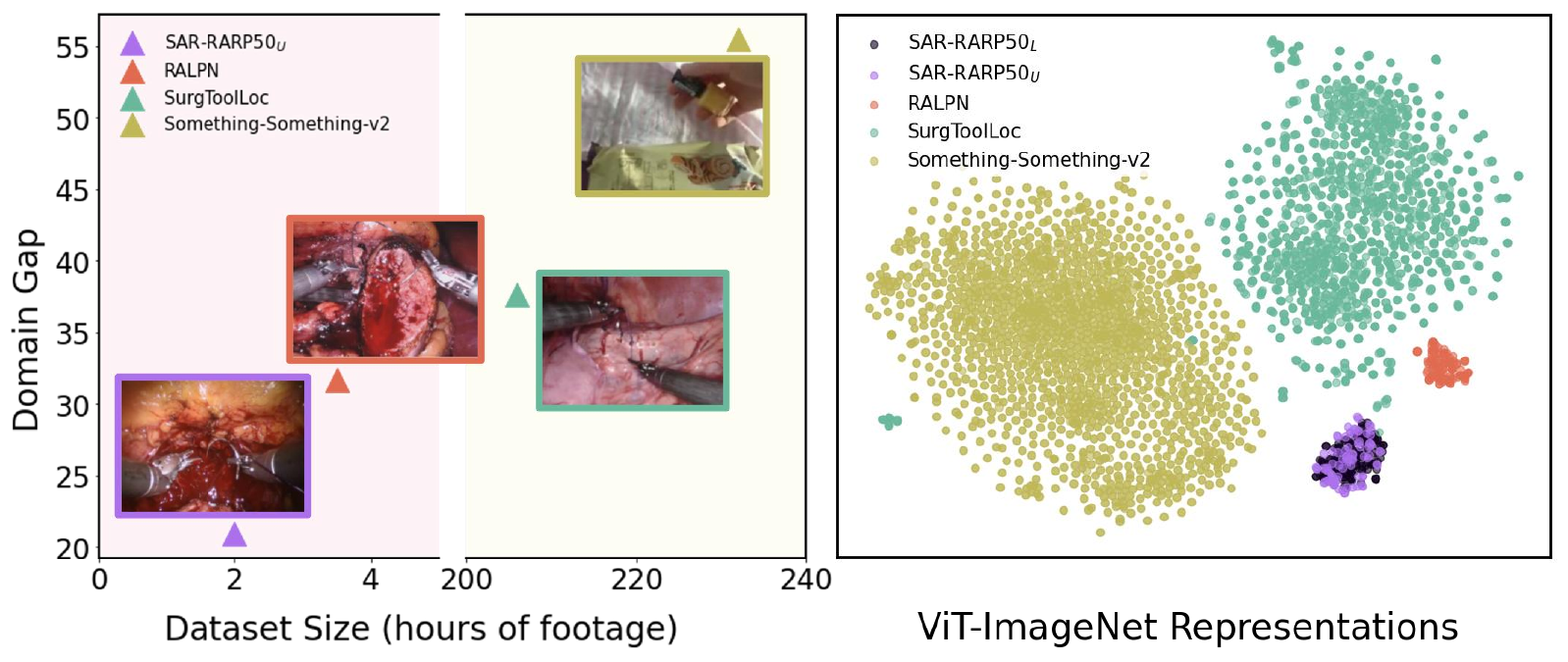}
    \caption{Left: EMD-based domain gap of each dataset with respect to SAR-RARP50\textsubscript{L} plotted against dataset size (in hours of footage). Right: t-SNE visualization of ViT-ImageNet representations.
} \label{fig_domain_gap}
   
\end{figure*}

\subsection{Domain Gap}
We propose estimating the \textit{domain gap} between each pre-training dataset and the downstream dataset using the Earth Mover’s Distance (EMD) metric \cite{Rubner1998,Oh2022}. We uniformly sample $K$ non-overlapping 16-frame snippets from each video. A video $v_i$ is thus represented as a tensor $v_i \in \mathbb{R}^{K \times 16 \times c \times h \times w}$, where $c$, $h$, and $w$ are the number of channels, height, and width of each frame, respectively. Given a spatial feature extractor $f: \mathbb{R}^{c \times h \times w} \rightarrow \mathbb{R}^{d}$, we extract features from each frame in the video to obtain a feature matrix $F_i \in \mathbb{R}^{K \times 16 \times d}$, where $d$ is the dimensionality of the extracted features. We then average over the temporal (16-frame) dimension to obtain snippet-level features, resulting in ${F}_i \in \mathbb{R}^{K \times d}$ for each video. Applying this process to all videos in a dataset yields a distribution of features $F^{\text{dist}} \in \mathbb{R}^{N \cdot K \times d}$, where $N$ is the number of videos in the dataset. Finally, the domain gap between two datasets is defined as the EMD computed between their respective feature distributions. For feature extraction, we utilize a Vision Transformer \cite{Dosovitskiy2021} encoder pre-trained on ImageNet.

Fig.~\ref{fig_domain_gap} summarizes our analysis. The left panel shows the domain gap plotted against the dataset size (in hours of footage), while the right panel presents the corresponding t-SNE visualizations of the feature embeddings. The datasets are ranked by their domain gap to SAR-RARP50\textsubscript{L} as follows: SAR-RARP50\textsubscript{U} $<$ RALPN $<$ SurgToolLoc $<$ Something-Something-v2.

\section{Experiments}
\subsection{Training Procedure and Evaluation Protocol}

\begin{figure}[t]
    \centering
    \includegraphics[width=1\textwidth]{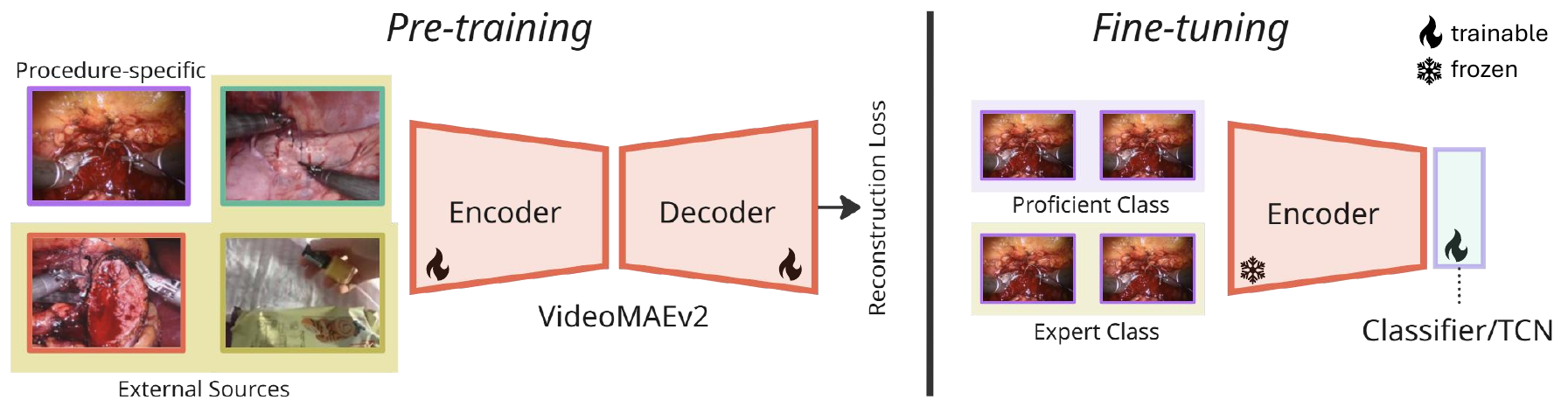}
    \caption{Overview of our investigation framework: Pre-training is performed using the VideoMAEv2 model on various individual and combined data sources. Fine-tuning is then conducted on the few-shot SSA task, where the frozen encoder is paired with a trainable classifier or TCN.
} \label{fig_method}
   
\end{figure}

The pre-training and evaluation framework is shown in Fig.~\ref{fig_method}. The SSL framework we adopt for pre-training is VideoMAEv2 \cite{Wang2023}, a video-based masked autoencoder that has demonstrated strong performance in action recognition tasks. We used the ViT-Small backbone with a 4-layer decoder, sampling 16-frame snippets, and trained with the AdamW optimizer for 300 epochs, batch size of 96, and learning rate of 0.0006.

After pre-training, we freeze the encoder and extract spatio-temporal features from 16-frame clips of the downstream dataset. We then fine-tune using the AdamW optimizer for 30 epochs with a learning rate of 0.001, a cosine learning rate scheduler, and a batch size of 1. To assess the transferability of the learned representations, we conduct two types of downstream evaluations: (1) Linear evaluation, where a linear classifier is trained on top of the frozen encoder features; (2) Temporal evaluation, where a Temporal Convolutional Network (TCN) \cite{Anastasiou2023} is applied to capture temporal dynamics over the extracted features. Few-shot performance is evaluated in the 1-shot, 2-shot, and 5-shot settings, averaged over 100 randomly sampled episodes, following standard few-shot evaluation protocols~\cite{Oh2022,Dhillon2020}. We report the average classification Accuracy and the F1-score, and their corresponding standard deviations. All experiments were implemented in PyTorch and carried out on NVIDIA RTX A6000 GPUs.

\subsection{Pre-training Configurations}
We design two types of pre-training configurations:
\begin{enumerate}
    \item \textbf{Single-dataset pre-training}: where the model is pre-trained using only one dataset. This configuration investigates the effect of domain gap on the downstream SSA task.
    \item  \textbf{Combined-dataset pre-training}: where we pre-train on SAR-RARP50\textsubscript{U} together with one additional dataset. This setup allows us to examine whether incorporating procedure-specific unlabeled data in conjunction with external datasets can enhance the quality of learned representations.
\end{enumerate}
We also report results for a randomly initialized encoder, and EndoViT \cite{Batic2024}, a recent surgical foundation model, as baselines.

\section{Results and Discussion}
\begin{figure}[t]
    \centering
    \includegraphics[width=1\textwidth]{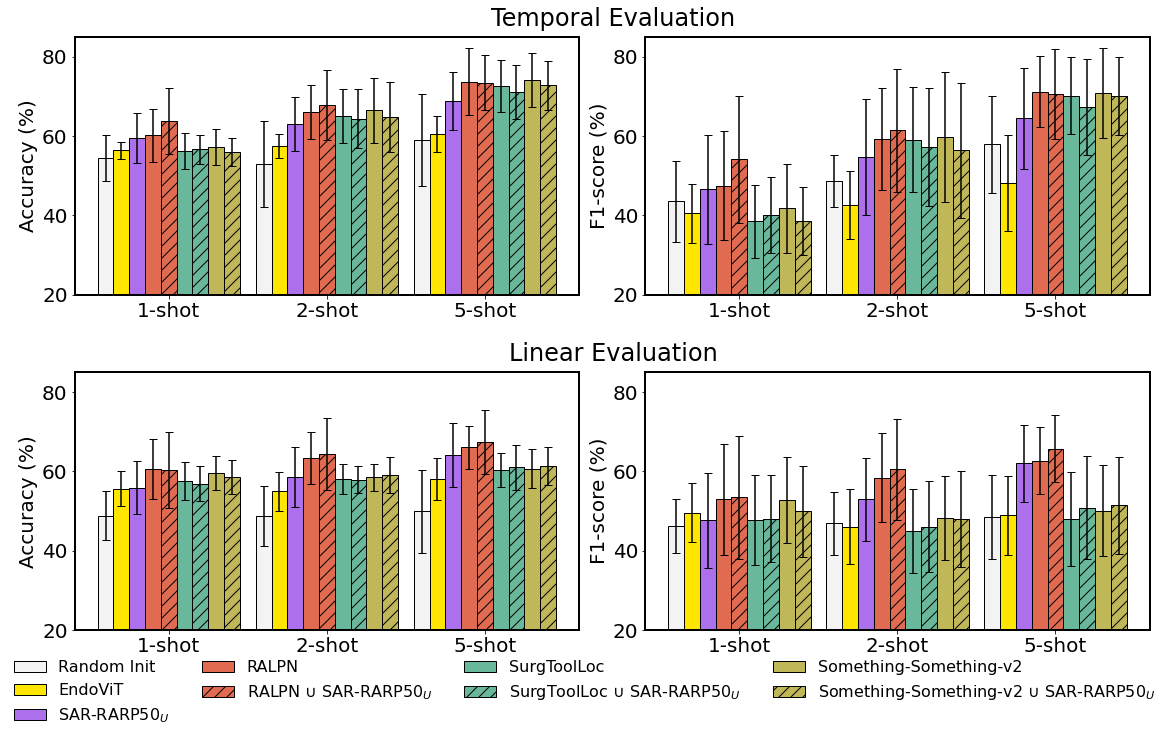}
    \caption{Classification Accuracy and F1-score for each pre-training configuration evaluated on the downstream SSA task across 1-shot, 2-shot, and 5-shot settings, and, temporal (top) and linear (bottom) evaluation.
} \label{fig_results}
   
\end{figure}

Fig.~\ref{fig_results} presents the performance of the the pre-training configurations on the downstream task of SSA, evaluated across three \textit{k}-shot settings and two evaluation setups. In the temporal evaluation, RALPN consistently demonstrates strong performance across all shot settings. In the 1-shot setting, it achieves an accuracy of 60.16\% slightly outperforming the procedure-specific SAR-RARP50\textsubscript{U} pre-training with 59.48\% accuracy. As the number of shots increases, the large-scale datasets SurgToolLoc and Something-Something-v2 begin to close the performance gap or even slightly outperform RALPN. In the 5-shot setting, Something-Something-v2 achieves the highest accuracy and F1-score with 74.09\% and 70.9\%, respectively, followed by SurgToolLoc (72.61\%, 70.2\%) and RALPN (73.65\%, 71.2\%). These results indicate that pre-training on large-scale datasets can be particularly beneficial for temporal models when more labeled data is available during fine-tuning.

However, the trend differs in the linear evaluation setting. While RALPN again achieves the highest performance across all shots, both SurgToolLoc and Something-Something-v2 perform considerably worse, particularly in terms of F1-score. In the 1-shot setting, their accuracies remain around 57.55\%–59.55\%, with corresponding F1-scores between 47.7\% and 52.7\%. Even as the number of shots increases, their performance lags significantly behind RALPN. This suggests that the large domain gap results in less effective representations from these datasets, which may lack the task-specific discriminative features necessary for effective linear classification in few-shot SSA.

Interestingly, despite its relatively small size, RALPN consistently delivers robust results, which implies that highly relevant data, even in small quantities, can be more effective for few-shot SSA than large-scale but less domain-aligned datasets. This is further supported by the observation that SAR-RARP50\textsubscript{U}, a procedure-specific pre-training set, also performs competitively, while slightly worse than RALPN, potentially due to the latter's larger size. Furthermore, although SurgToolLoc exhibits a smaller domain gap to SAR-RARP50\textsubscript{L} compared to Something-Something-v2, the latter slightly outperforms it in most settings, suggesting that pre-training on SurgToolLoc does not yield sufficiently relevant or discriminative features to benefit the downstream SSA task.

\definecolor{darkgreen}{RGB}{13, 138, 71}

\begin{table}[t]
\centering
\caption{Average performance gain (\%) for combined-dataset pre-training configurations compared to individual datasets. Gains are reported for both accuracy and F1-score, averaged across all shot settings and evaluation types.}
\label{table_comparison}
\resizebox{\textwidth}{!}{%
\begin{tabular}{l|cccccc}
\hline
\multirow{3}{*}{Performance Gain (\%)} & \multicolumn{6}{c}{Pre-training Configurations Comparison}                                                                                                                                                                                                                                             \\ \cline{2-7} 
                                            & \multicolumn{2}{c|}{\begin{tabular}[c]{@{}c@{}}RALPN$\cup$SAR-RARP50\textsubscript{U}\\ vs\end{tabular}} & \multicolumn{2}{c|}{\begin{tabular}[c]{@{}c@{}}Something-Something-v2$\cup$SAR-RARP50\textsubscript{U}\\ vs\end{tabular}} & \multicolumn{2}{c}{\begin{tabular}[c]{@{}c@{}}SurgToolLoc$\cup$SAR-RARP50\textsubscript{U}\\ vs\end{tabular}} \\ \cline{2-7} 
                                            & \multicolumn{1}{c|}{RALPN}            & \multicolumn{1}{c|}{SAR-RARP50\textsubscript{U}}            & \multicolumn{1}{c|}{Something-Something-v2}            & \multicolumn{1}{c|}{SAR-RARP50\textsubscript{U}}            & \multicolumn{1}{c|}{SurgToolLoc}                       & SAR-RARP50\textsubscript{U}                      \\ \hline
Avg. Accuracy Gain                          & \multicolumn{1}{c|}{\textcolor{darkgreen}{+1.22}}            & \multicolumn{1}{c|}{\textcolor{darkgreen}{+4.56}}                 & \multicolumn{1}{c|}{\textcolor{red}{-0.65}}                             & \multicolumn{1}{c|}{\textcolor{darkgreen}{+0.46}}                 & \multicolumn{1}{c|}{\textcolor{red}{-0.35}}                             & \textcolor{darkgreen}{+0.18}                           \\
Avg. F1-score Gain                          & \multicolumn{1}{c|}{\textcolor{darkgreen}{+2.28}}            & \multicolumn{1}{c|}{\textcolor{darkgreen}{+6.2}}                  & \multicolumn{1}{c|}{\textcolor{red}{-1.53}}                             & \multicolumn{1}{c|}{\textcolor{red}{-2.35}}                 & \multicolumn{1}{c|}{\textcolor{red}{-0.34}}                             & \textcolor{red}{-3.15}                           \\ \hline
\end{tabular}%
}
\end{table}

Furthermore, EndoViT, despite being pre-trained on Endo700k, a large-scale collection of nine diverse endoscopic datasets, underperforms across all settings and, in some cases, even falls below random initialization. This suggests that excessive diversity across surgical procedures and anatomical contexts may dilute the relevance of the learned representations for downstream tasks such as SSA, which require fine-grained procedural understanding. However, this is not a fully direct comparison, as EndoViT uses the Masked Autoencoder (MAE) framework \cite{He2022} that focuses solely on spatial reconstruction, whereas VideoMAEv2 incorporates temporal reconstruction, which may be more suitable for capturing the dynamic cues necessary for SSA.

For the combined-dataset pre-training configurations, where procedure-specific data is merged with external datasets, we compute the average performance gain in accuracy and F1-score across all shot settings and evaluation types. The results, presented in Table~\ref{table_comparison}, show that combining RALPN with SAR-RARP50\textsubscript{U} consistently improves performance compared to using either dataset for pre-training alone. This suggests a strong alignment between the two datasets, likely due to their shared domain characteristics. In contrast, combining SAR-RARP50\textsubscript{U} with more dissimilar external datasets Something-Something-v2 or SurgToolLoc leads to inconsistent outcomes. In some cases, performance improves slightly, but in others, it degrades, suggesting that a large domain gap can introduce noise that hinders effective pre-training. These findings suggest that integrating procedure-specific data with external sources is only beneficial for downstream SSA when the domain gap between datasets is small. When the gap is large, such combinations may offer little advantage or even harm performance. A table with side-by-side comparisons of the results presented in Fig.~\ref{fig_results} can be found in the Supplementary material.


\section{Conclusions and Future Work}
We presented a preliminary investigation into the impact of self-supervised pre-training strategies on few-shot SSA, a task that remains underexplored despite its clinical significance. By evaluating pre-training datasets of varying sizes and domain similarity to a real-world robotic surgery dataset, we examined how domain gap influences downstream performance in few-shot settings. Our findings show that small but domain-relevant datasets can outperform much larger, less aligned ones, even when the latter are also surgical. Furthermore, adding procedure-specific data to a related external dataset during pre-training significantly boosts performance, whereas combining procedure-specific data with less similar but larger-scale sources degrades it. These insights highlight the importance of domain alignment in pre-training design and provide a foundation for a new and promising direction of research in SSA.

While our findings are promising, they also present opportunities for future work. First, we limited our evaluation to a single real-world surgical dataset due to the high computational demands of video-based SSL. Extending the analysis to additional datasets from different centers would be valuable to validate the generalizability of our conclusions. Second, our formulation simplifies SSA into a binary classification task. This decision was driven by the limited range of GRS scores in the dataset, making fine-grained classification infeasible due to severe class imbalance. Expanding the number of annotated samples across the score range could allow for more granular skill categories in future work. Lastly, although we focused on SSA due to its clinical relevance and complexity, our experimental framework could be further evaluated in other video-level surgical tasks that involve one label per video.

\begin{credits}
\subsubsection{\ackname} 
This work was supported by the EPSRC under the UCL Doctoral Training Partnership (DTP) [EP/R513143/1, EP/T517793/1], the UCL Centre for Doctoral Training in Intelligent, Integrated Imaging in Healthcare (i4health) [EP/S021930/1] and the Human-centred Machine Intelligence to optimise Robotic Surgical Training (HuMIRoS) project [EP/Z534754/1]; the Optical and Acoustic imaging for Surgical and Interventional Sciences (OASIS) project [UKRI145]; the Department of Science, Innovation and Technology (DSIT) and the Royal Academy of Engineering under the Chair in Emerging Technologies programme; and the Wolfson Foundation. For the purpose of open access, the author has applied a CC BY public copyright licence to any author accepted manuscript version arising from this submission.

\subsubsection{\discintname}
The authors have no competing interests to declare that are relevant to the content of this article.
\end{credits}
%
%
%
\bibliographystyle{splncs04}
\bibliography{demi}

\end{document}